\documentclass{llncs}

\usepackage{graphicx,graphics}
\usepackage{color}
\usepackage{url}
\usepackage{listings}
\usepackage{subfigure}

\lstdefinelanguage{Comet}
	{ morekeywords={range, enum, int, inc, var, forall, in, while,
	do, if, else, return, void, set, float, Float, move, sum,
	prod, class, tuple, interface, function, operator, implements,
	extends, import, include, native, this, super, union, cross,
	and, or, union, inter, string, Closure, Counter, closure,
	pclosure, Process, break, continue, select, selectMin,
	selectMinAsp, selectMax, selectMaxAsp, selectFirst,
	selectRandom, selectCircular, selectPr, maximizing,
	minimizing, minimize, maximize, minmax, max, argMax, min,
	argMin, new, expr, setof, collect, filter, mapof, all, abs,
	floor, ceil, sqrt, ln, lookahead, IntToString, card, when,
	whenever, foreveron, sleepUntil, neighbor, Boolean, Integer,
	abstract, continuation,
	Continuation,Solution,CPNode,Checkpoint,model, soft, hard,
	call, instanceof, try, tryall, parall, pardo, until, parever,
	onFailure, exploreall, explore, catch, throw, cast, ref, sync,
	shared, dict, implements, switch, case, default, null, Object,
	boolean, bool, true, false, Event, KeyEvent, Condition,
	notify, clearEvents, with, interval, assert, member, thread,
	for, synchronized, by, final, noevent, alias, trail, subject,
	to, using, sin, cos, atan, ln, exp,onDomains,onValues},
	  sensitive=true,
	  basicstyle=\footnotesize,
	  tabsize=4,
	  frame=lines,
	  numbers=left,
	  stepnumber=1,
	  numberstyle=\tiny,
	  numbersep=3pt,	
	  xleftmargin=10pt,
	  columns=fullflexible,
	  morecomment=[l]{//},
	  morecomment=[s]{/*}{*/},
	  morestring=[b]",
	  morestring=[d]',
	  showstringspaces=false
	}
	
\lstnewenvironment{Comet}{\lstset{language=Comet}}{}

\newcommand{\remove}[1]{}

\newcommand{\vsids}{\textsc{Vsids}}
\newcommand{\ibs}{\textsc{Ibs}}
\newcommand{\abs}{\textsc{Abs}}
\newcommand{\wdeg}{\textsc{wdeg}}

\newcommand{\elsdel}[1]{{\color{red}}}

\begin{document}

\title{Activity-Based Search \\
  for Black-Box Contraint-Programming Solvers }

\titlerunning{Activity-based Search for Black-Box Contraint-Programming Solvers}

\author{ L. Michel\inst{2} \and
P. Van Hentenryck\inst{1}
}

\institute{
 Brown University, Box 1910, Providence, RI 02912 \\
 \and
 University of Connecticut, Storrs, CT 06269-2155
}
\maketitle

\begin{abstract}
  Robust search procedures are a central component in the design of
  black-box constraint-programming solvers. This paper proposes
  activity-based search, the idea of using the activity of variables
  during propagation to guide the search. Activity-based search was
  compared experimentally to impact-based search and the \wdeg{}
  heuristics. Experimental results on a variety of benchmarks show
  that activity-based search is more robust than other heuristics and
  may produce significant improvements in performance.
\end{abstract}

\section{Introduction}

Historically, the constraint-programming (CP) community has focused on
developing open, extensible optimization tools, where the modeling and
the search procedure can be specialized to the problem at hand. This
focus stems partially from the roots of CP in programming languages
and partly from the rich modeling language typically found in CP
systems. While this flexibility is appealing for experts in the field,
it places significant burden on practitioners, reducing its acceptance
across the wide spectrum of potential users. In recent years however,
the constraint-programming community devoted increasing
attention to the development of black-box constraint solvers. This new
focus was motivated by the success of Mixed-Integer Programming (MIP)
and SAT solvers on a variety of problem classes.  MIP and SAT solvers
are typically black-box systems with automatic model reformulations
and general-purpose search procedures. As such, they allow
practitioners to focus on modeling aspects and may reduce the time to
solution significantly.

This research is concerned with one important aspect of black-box
solvers: the implementation of a robust search procedure. In recent
years, various proposals have addressed this issue.  Impact-based
search (\ibs{}) \cite{Refalo04} is motivated by concepts found in MIP
solvers such as strong branching and pseudo costs. Subsequent work
about solution counting can be seen as an alternative to impacts
\cite{Pesant09} that exploits the structure of CP constraints. The
weighted-degree heuristic (\wdeg{}) \cite{BoussemartHLS04} is a direct
adaptation of the SAT heuristic \vsids{}\cite{Moskewicz:2001} to CSPs
that relies on information collected from failures to define the
variable ordering.

This paper proposes Activity-Based Search (\abs{}), a search heuristic
that recognizes the central role of constraint propagation in
constraint-programming systems. Its key idea is to associate with each
variable a counter which measures the activity of a variable during
propagation. This measure is updated systematically during search and
initialized by a probing process. \abs{} has a number of advantages
compared to earlier proposals. First, it does not deal explicitly with
variable domains which complicates the implementation and runtime
requirements of \ibs{}. Second, it does not instrument constraints
which is a significant burden in solution-counting heuristics. Third,
it naturally deals with global constraints, which is not the case of
\wdeg{} since all variables in a failed constraint receive the same
weight contribution although only a subset of them is relevant to the
conflict. \abs{} was compared experimentally to \ibs{} and \wdeg{} on
a variety of benchmarks. The results show that \abs{} is the most
robust heuristic and can produce significant improvements in
performance over \ibs{} and \wdeg{}, especially when the problem
complexity increases.

The rest of the paper is organized as follows.  Sections
\ref{sec:impact} and \ref{sec:wdeg} review the \ibs{} and \wdeg{}
heuristics.  Section \ref{sec:activity} presents \abs{}. Section
\ref{sec:experimental} presents the experimental results and Section
\ref{sec:ccl} concludes the paper.

\section{Impact-Based Search}
\label{sec:impact}

Impact-based search was motivated by the concept of pseudo-cost in MIP
solvers.  The idea is to associate with a branching decision $x=v$ a
measure of how effectively it shrinks the search space. This measure
is called the \textit{impact} of the branching decision. 

\paragraph{Formalization}

Let $P=\langle X,D,C\rangle$ be a CSP defined over variables $X$,
domains $D$, and constraints $C$. Let $D(x_i)$ denote the domain of
variable $x_i \in X$ and $|D(x_i)|$ denote the size of this domain.  A
trivial upper-bound on the size of the search space of ${\cal S}(P)$
is given by the product of the domain sizes:
\[
{\cal S}(P) = \prod_{x \in X} |D(x)|
\]
At node $k$, the search procedure receives a CSP $P_{k-1}=\langle
X,D_{k-1},C_{k-1}\rangle$, where $C_{k-1} = C
\cup\{c_0,c_1,c_2,\cdots,c_{k-1}\}$ and $c_i$ is the constraint posted
at node $i$. Labeling a variable $x$ with value $a \in D_{k-1}(x)$
adds a constraint $x=a$ to $C_{k-1}$ to produce, after propagation,
the CSP $P_{k}=\langle X,D_{k},C_{k}\rangle$.

The contraction of the search space induced by a labeling $x=a$ is
defined as
\[
I(x=a) = 1 - \frac{{\cal S}(P_k)}{{\cal S}(P_{k-1})}
\]
$I(x=a)=1$ when the assignment produces a failure since ${\cal S}(P_k)
= 0$ and $I(x=a) \approx 0$ whenever ${\cal S}(P_k) \approx {\cal
  S}(P_{k-1})$, i.e., whenever there is almost no domain reduction.
An \textit{estimate} of the impact of the labeling constraint $x=a$
over a set of search tree nodes ${\cal K}$ can then be defined as
\[
\bar{I}(x=a) = \frac{\sum_{k \in {\cal K}} 1 - \frac{{\cal S}(P_k)}{{\cal S}(P_{k-1})}}{|\cal K|}
\]
\noindent Actual implementations (e.g., \cite{googleSolver10}) rely
instead on
\[
\bar{I}_1(x=a) = \frac{\bar{I}_0(x=a) \cdot (\alpha - 1) + I(x=a)}{\alpha}
\]
\noindent 
where $\alpha$ is a parameter of the engine and the subscripts in
$\bar{I}_0$ and $\bar{I}_1$ denote the impact before and after the
update. Clearly, $\alpha=1$ yields a forgetful strategy, $\alpha=2$
gives a running average that progressively decays past impacts, while
a choice $\alpha > 2$ favors past information over most recent
observations.

The (approximate) impact of a variable $x$ at node $k$ is defined as
\[
{\cal I}(x) = \sum_{a \in D_k(x)}  1 - \bar{I}(x=a)
\]
To obtain suitable estimates of the assignment and variable impacts at
the root node, \ibs{} simulates all the $\sum_{x \in X} |D(x)|$
possible assignments. For large domains, domain values are partitioned
in blocks. Namely, for a variable $x$, let $D(x) = \cup_{i=1}^b B_i$
with $B_i \cap B_j = \emptyset \:\: \forall i,j : i\neq j \in
1..b$. The impact of a value $a \in B_i$ ($i \in 1..b$) is then set to
$I(x=a) = I(x \in B_i)$. With partitioning, the initialization costs
drop from $|D(x)|$ propagations to $b$ propagations (one per block).
The space requirement for \ibs{} is $\Theta(\sum_{x \in X} |D(x)|)$,
since it stores the impacts of all variable/value pairs.

\paragraph{The Search Procedure}

\ibs{} defines a variable and a value selection heuristic.  \ibs{}
first selects a variable $x$ with the largest impact, i.e., $x \in
\mbox{argMax}_{x \in X} {{\cal I}(x)}$. It then selects a value $a$
with the least impact, i.e., $a \in \mbox{argMin}_{v \in D(x)}
\bar{I}(x=v)$.  Neither $\mbox{argMax}_{x \in X} {{\cal I}(x)}$ nor
$\mbox{argMin}_{v \in D(x)} \bar{I}(x=v)$ are guaranteed to be a
singleton and, in case of ties, \ibs{} breaks the ties uniformly at
random.

As any randomized search procedure, \ibs{} can be augmented with a
restart strategy. A simple restarting scheme limits the number of
failures in round $i$ to $l_i$ and increases the limit between rounds
to $l_{i+1} = \rho \cdot l_i$ where $\rho > 1$.

\section{The \textsc{WDEG} Heuristic}
\label{sec:wdeg}

\wdeg{} maintains, for each constraint, a counter (weight)
representing the number of times a variable appears in a failed
constraint, i.e., a constraint whose propagation removes all values in
the domain of a variable. The weighted degree of variable $x$ is
defined as
\[
\alpha_{wdeg}(x) = \sum_{c \in C} weight[c] \mbox{ s.t. } x \in vars(c) x \wedge |FutVars(c)|>1
\]
where $FutVars(c)$ is the set of uninstantiated variables in $c$.

\wdeg{} only defines a variable selection heuristic: It first selects
a variable $x$ with the smallest ratio
$\frac{|D(x)|}{\alpha_{wdeg}(x)}$. All the weights are initialized
to 1 and, when a constraint fails, its weight is incremented. The
space overhead of \wdeg{} is $\Theta(|C|)$ for a CSP $\langle
X,D,C\rangle$.

\section{Activity-Based Search}
\label{sec:activity}

\abs{} is motivated by the key role of propagation in
constraint-programming solvers. Contrary to SAT solvers, CP uses
sophisticated filtering algorithms to prune the search space by
removing values that cannot appear in  solutions.  \abs{} exploits
this filtering information and maintains, for each variable $x$, a
measure of how often the domain of $x$ is reduced during the
search. The space requirement for this statistic is $\Theta(|X|)$.
\abs{} can {\em optionally} maintain a measure of how much activity
can be imputed to each assignments $x=a$ in order to drive a
value-selection heuristic. If such a measure is maintained, the space
requirement is proportional to the number of distinct assignments
performed during the search and is bounded by ${\cal O}(\sum_{x \in X}
|D(x)|)$. \abs{} relies on a decaying sum to forget the oldest
statistics progressively, using an idea from \vsids{}. It also
initializes the activity of the variables by probing the search space.

\abs{} is simple to implement and does not require sophisticated
constraint instrumentation. It scales to large domains without special
treatment and is independent of the domain sizes when the value
heuristic is not used. Also, \abs{} does not favor variables appearing in
failed constraints, since a failure in a CP system is typically the
consequence of many filtering algorithms.

\paragraph{Formalization}

Given a CSP $P=\langle X,D,C\rangle$, a CP solver applies a
constraint-propagation algorithm $F$ after a labeling decision.  $F$
produces a new domain store $D' \subseteq D$ enforcing the required
level of consistency. Applying $F$ to $P$ identifies a subset $X'
\subseteq X$ of affected variables defined by
\[
\begin{array}{ll}
\forall x \in X' &: D'(x) \subset D(x);   \\
\forall x \in X \setminus X' &: D'(x) = D(x).
\end{array}
\]
The {\em activity} of $x$, denoted by $A(x)$, is updated at each node $k$
of the search tree by the following two rules:
\[
\begin{array}{lcl}
\forall x \in X  \mbox{ s.t. } |D(x)| > 1 &:& A(x) = A(x) \cdot \gamma \\
\forall x \in X' &:& A(x) = A(x) + 1
\end{array}
\]
where $X'$ is the subset of affected variables and $\gamma$ be an age
decay parameter satisfying $0 \leq \gamma \leq 1$.  The aging only
affects free variables since otherwise it would quickly erase the
activity of variables labeled early in the search.

The activity of an assignment $x=a$ at a search node $k$ is defined
as the number of affected variables in $|X'|$ when applying $F$ on $C
\cup \{x=a\}$, i.e.,
\[
A_k(x=a) = |X'|.
\]
As for impacts, the activity of $x=a$ over the entire tree can be
estimated by an average over all the tree nodes seen so far, i.e.,
over the set of nodes ${\cal K}$. The estimation is thus defined as
\[
\tilde{A}(x=a) = \frac{\sum_{k \in {\cal K}} A_k(x=a)}{|{\cal K}|}
\]
Once again, it is simpler to favor a weighted sum instead
\[
\tilde{A}_1(x=a)  = \frac{\tilde{A}_0(x=a) \cdot (\alpha - 1) + A_k(x=a)}{\alpha}
\]
where the subscripts on $\tilde{A}$ capture the estimate before and
after the update. When the value heuristic is not used, it is not
necessary to maintain $\tilde{A}(x=a)$ {\em which reduces the space
  requirements  without affecting variable activities.}

\paragraph{The Search Procedure}

\abs{} defines a variable ordering and possibly a value ordering. It
selects the variable $x$ with the largest ratio $\frac{A(x)}{|D(x)|}$,
i.e., the most active variable per domain value. Ties are broken
uniformly at random. When a value heuristic is used, \abs{} selects a
value $a$ with the least activity, i.e., $a \in \mbox{argMin}_{v \in
  D(x)} \tilde{A}(x=v)$.  The search procedure can be augmented with
restarts. The activities can be used ``as-is'' to guide the search
after a restart. It is also possible to reinitialize activities in
various ways, but this option was not explored so far in the
experimentations.

\paragraph{Initializing Activities}

\abs{} uses probing to initialize the activities. Consider a path
$\pi$ going from the root to a leaf node $k$ in a search tree for the
CSP $P=\langle X,D,C\rangle$. This path $\pi$ corresponds to a
sequence of labeling decisions $(x_0=v_0, x_1=v_1, \cdots, x_k=v_k)$
in which the $j^{th}$ decision labels variable $x_j$ with $v_j \in
D_j(x_j)$. If $X_{j} \subseteq X$ is the subset of variables whose
domains are filtered as a result of applying $F$ after decision $x_j =
v_j$, the activity of variable $x$ along path $\pi$ is defined as
$A^{\pi}(x) = {A}^{\pi}_k(x)$ where
\[
\left\{\begin{array}{lclll}
{A}_0^{\pi}(x) &=& 0  &\\
{A}_j^{\pi}(x) &=& {A}_{j-1}^{\pi}(x) + 1  &\Leftrightarrow x \in X_{j} & (1 \leq j \leq k) \\
{A}_j^{\pi}(x) &=& {A}_{j-1}^{\pi}(x) &\Leftrightarrow x \notin X_{j} & (1 \leq j \leq k) 
\end{array}
\right.
\]
${A}^{\pi}(x)=0$ if $x$ was never involved in any propagation
along $\pi$ and ${A}^{\pi}(x)=k$ if the domain of $x$ was
filtered by each labeling decision in $\pi$. Also, ${A}^{\pi}(x)
= A(x)$ when $\gamma=1$ (no aging) and path $\pi$ is followed.

Now let us now denote $\Pi$ the set of all paths in some search tree of
$P$.  Each such path $\pi \in \Pi$ defines an activity
${A}^{\pi}(x)$ for each variable $x$. Ideally, we would want to
initialize the activities of $x$ as the average over all paths in
$\Pi$, i.e.,
\[
\mu_{{A}}(x) = \frac{ \sum_{\pi \in \Pi} {A}^{\pi}(x) }{|\Pi|}.
\]   
\abs{} initializes the variables activities by sampling $\Pi$ to
obtain an estimate of the mean activity $\tilde{\mu_{{A}}}(x)$ from a
sample $\tilde{\Pi} \subset \Pi$. More precisely, \abs{} repeatedly
draws paths from $\Pi$. These paths are called \textit{probes} and the
$j^{th}$ assignment $x_j=v_j$ in a probe $p$ is selected uniformly at
random as follows: (1) $x_j$ is a free variable and (2) value $v_j$ is
picked from $D_j(x_j)$. During the probe execution, variable
activities are updated normally but no aging is applied in order to
ensure that all probes contribute equally to $\tilde{\mu_{{A}}}(x)$.
Observe that some probes may terminate prematurely since a failure may
be encountered; others may actually find a solution if they reach
a leaf node. Moreover, if if a failure is discovered at the root node,
singleton arc-consistency \cite{Prosser:2000} has been established and
the value is removed from the domain permanently.

The number of probes is chosen to provide a good estimate of the mean
activity over the paths. The probing process delivers an empirical
distribution $\tilde{A}(x)$ of the activity of each variable $x$ with
mean $\tilde{\mu_{{A}}}(x)$ and standard deviation
$\tilde{\sigma_{{A}}}(x)$. Since the probes are iid, the distribution
can be approximated by a normal distribution and the probing process
is terminated when the 95\% confidence interval of the t-distribution,
i.e., when
\[ [\tilde{\mu_{{A}}}(x) - t_{0.05,n-1} \cdot
\frac{\tilde{\sigma_{{A}}}(x)}{\sqrt{n}}, \tilde{\mu_{{A}}}(x) +
t_{0.05,n-1} \cdot \frac{\tilde{\sigma_{{A}}}(x)}{\sqrt{n}}]
\]
is sufficiently small (e.g., within $\delta\%$ of the empirical mean)
for all variables $x$ with $n$ being the number of probes,

Observe that this process does not require a separate
instrumentation. It uses the traditional activity machinery with
$\gamma=1$. In addition, the probing process does not add any space
requirement: the sample mean $\tilde{\mu_{{A}}}(x)$ and the sample
standard deviation $\tilde{\sigma_{{A}}}(x)$ are computed
incrementally, including the activity vector ${A}^p$ for each probe as
it is completed. If a value heuristic is used the sampling process also maintains $A(x=a)$
for every labeling decision $x=a$ attempted during the probes.

\section{Experimental Results}
\label{sec:experimental}

\subsection{The Experimental Setting}

\paragraph{The Configurations}

All the experiments were done on a Macbook Pro with a core i7 at
2.66Ghz running MacOS 10.6.7. \ibs{}, \wdeg{}, and \abs{} were were
all implemented in the \textsc{Comet} system \cite{Comet10}.  Since
the search algorithms are in general randomized, the empirical results
are based on 50 runs and the tables report the average ($\mu_T$) and
the standard deviation $\sigma_T$ of the running times in
seconds. Unless specified otherwise, a timeout of 5 minutes was used
and runs that timeout were assigned a $300s$ runtime. The following
parameter values were used for the experimental results: $\alpha=8$ in
both \ibs{} and \abs{}, $\gamma=0.999$ (slow aging), and
$\delta=20\%$. Experimental results on the sensitivities of these
parameters will also be reported. For every heuristic, the results are
given for three strategies: no restarts ($NR$), fast restarting
($\rho=1.1$) or slow restarting ($\rho=2$). depending on which
strategy is best across the board.  The initial failure limit is set
to $3 \cdot |X|$.

\paragraph{Search Algorithms}

The search algorithms were run on the exact same models, with a single
line changed to select the search procedure. In our experiments,
\ibs{} does \textit{not} partition the domains when initializing the
impacts and \textit{always} computes the impacts exactly. Both the
variable and value heuristics break ties randomly. In \wdeg{}, no
value heuristic is used: The values are tried in the sequential order
of the domain. Ties in the variable selection are broken randomly. All
the instances are solved using the same parameter values as
explained earlier. No comparison with model-counting heuristic is
provided, since these are not available in publicly available CP
solvers.

\paragraph{Benchmarks}

The experimental evaluation uses benchmarks that have been widely
studied, often by different communities. The multi-knapsack and magic
square problems both come from the \ibs{} paper \cite{Refalo04}. The
progressive party has been a standard benchmark in the local search,
mathematical-programming, and constraint-programming communities, and
captures a complex, multi-period allocation problem. The nurse
rostering problem \cite{Schaus09} originated from a
mathematical-programming paper and constraint programming was shown to
be a highly effective and scalable approach. The radiation problem is
taken from the 2008 MiniZinc challenge \cite{Nethercote07minizinc} and
has also been heavily studied. These benchmarks typically exploit many
features of constraint-programming systems including numerical,
logical, reified, element, and global constraints.

\subsection{The Core Results}

\paragraph{Multi-Knapsack}

\begin{figure}[t]
\includegraphics[width=0.5\columnwidth,trim=20 20 20 20,clip]{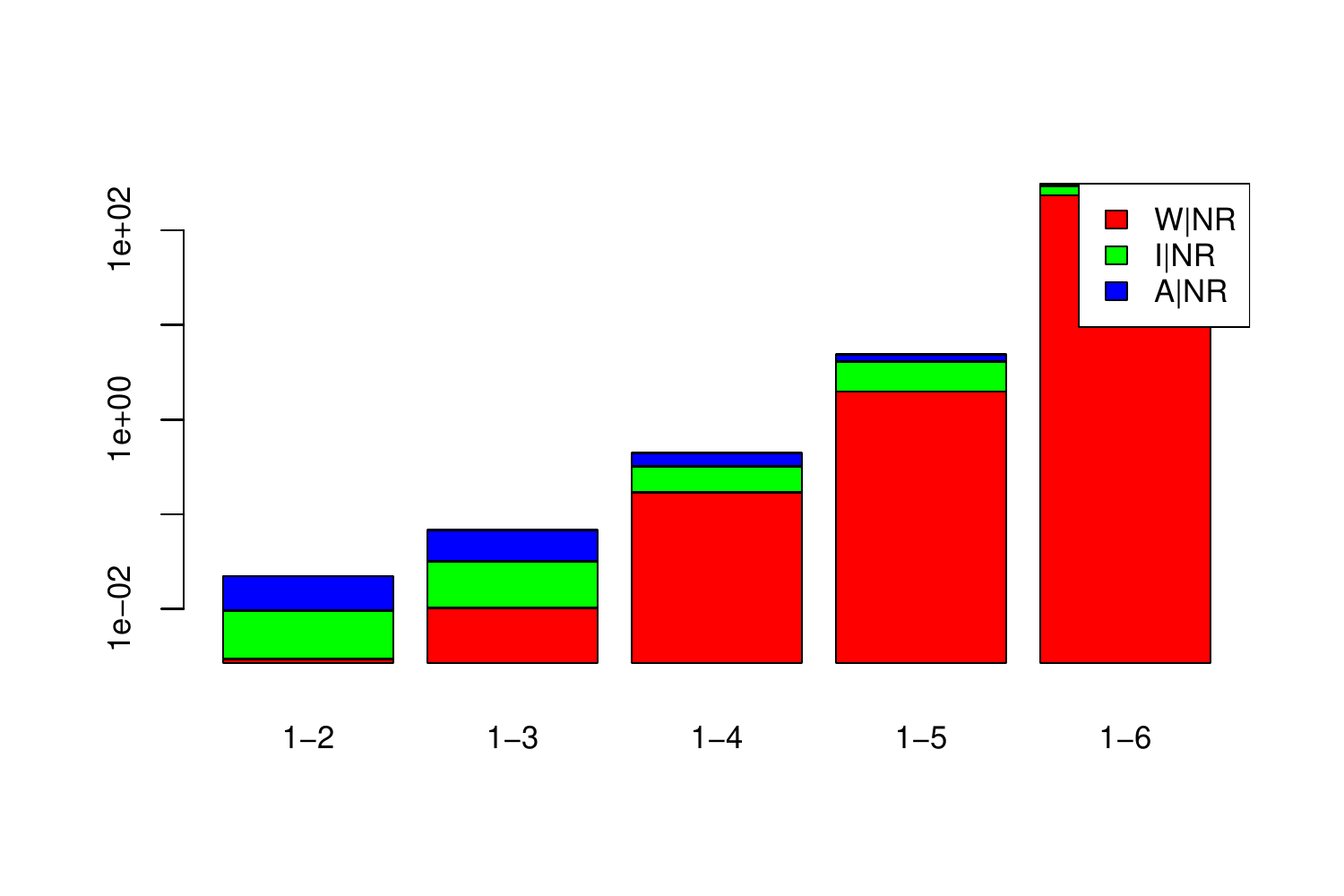}
\includegraphics[width=0.5\columnwidth,trim=20 20 20 20,clip]{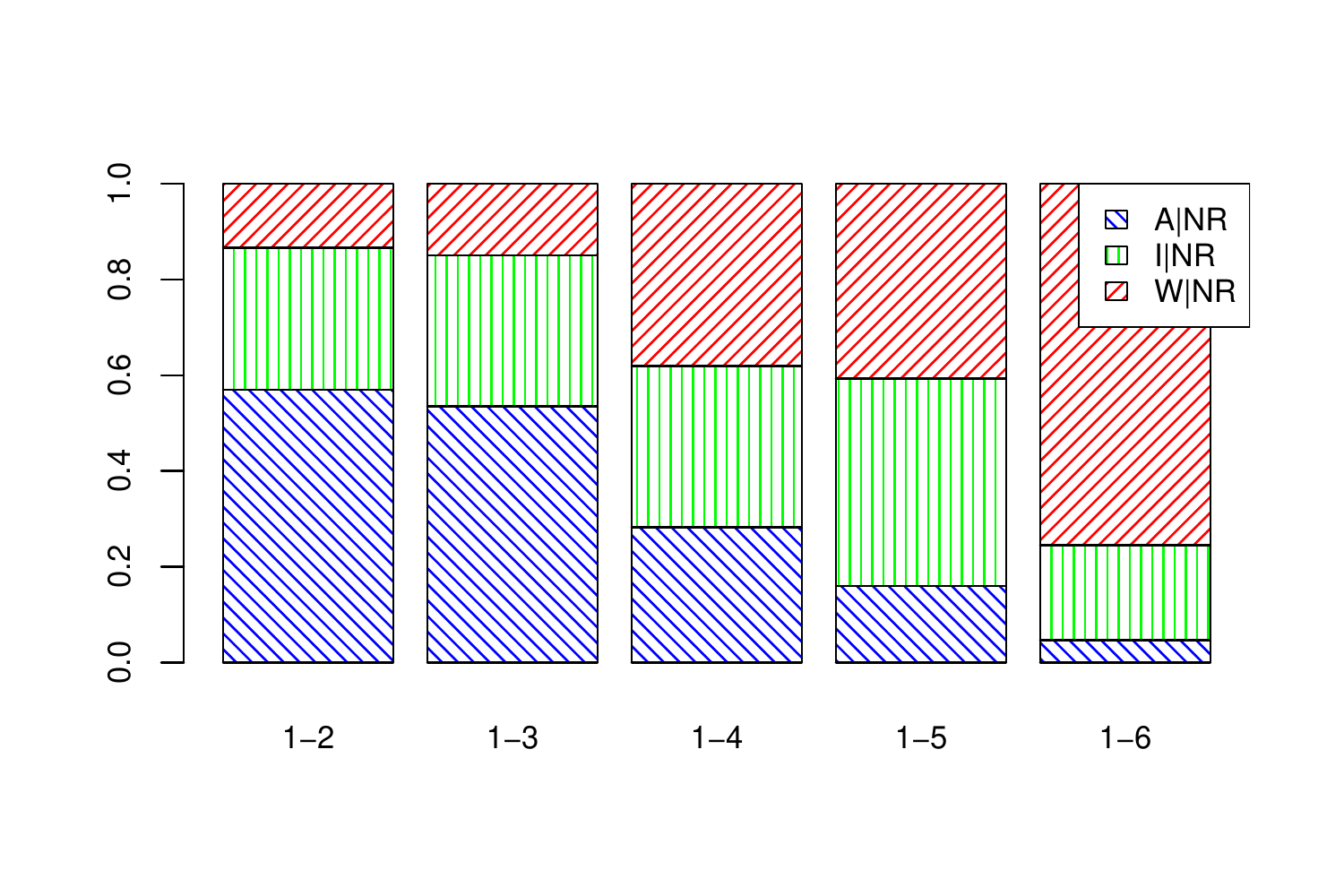}
\vspace{-10mm}
\caption{Knapsack, no-restart, decision-variant.}
\label{expe:csp-knap}
\end{figure}

\begin{figure}[t]
\includegraphics[width=0.5\columnwidth,trim=20 20 20 20,clip]{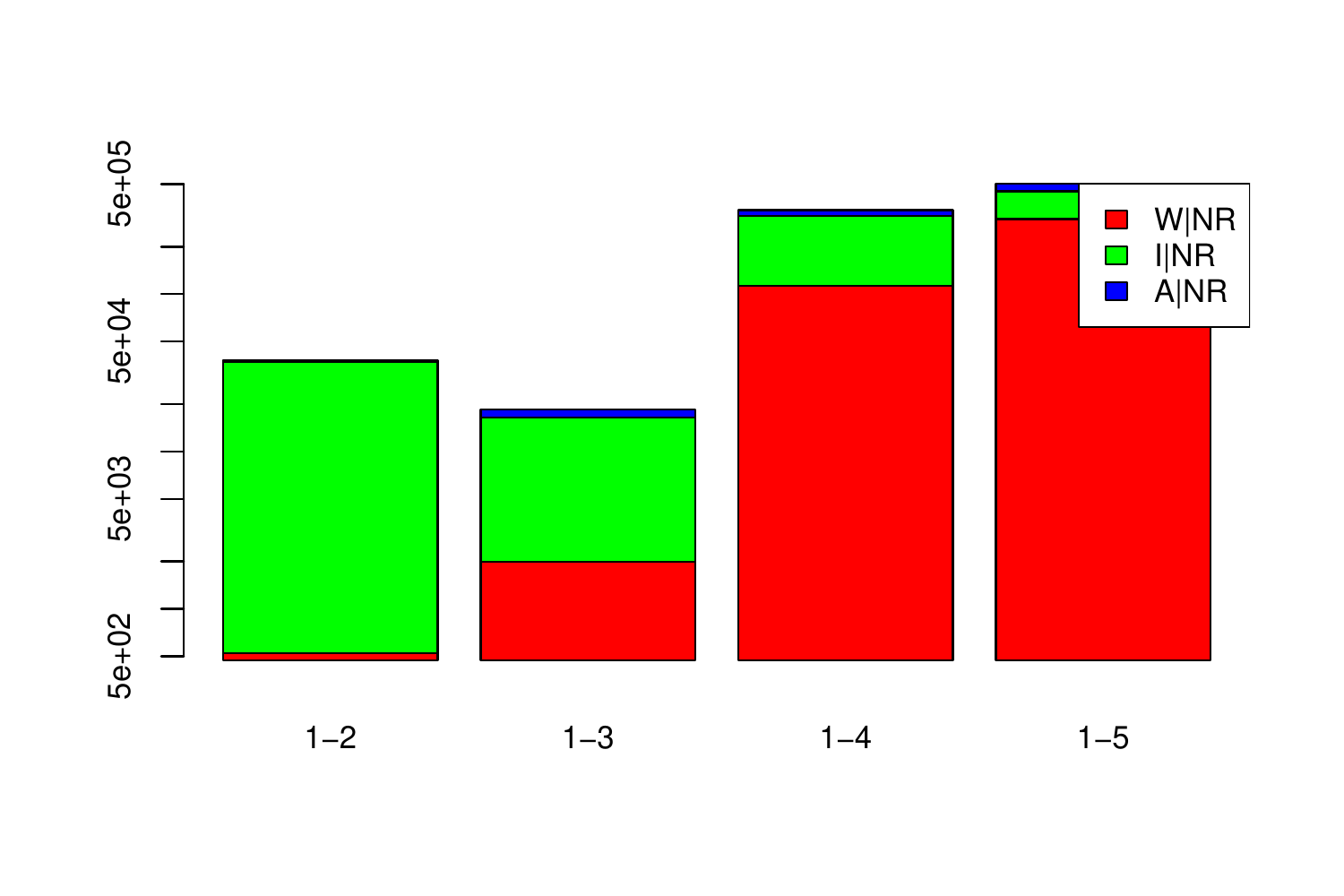}
\includegraphics[width=0.5\columnwidth,trim=20 20 20 20,clip]{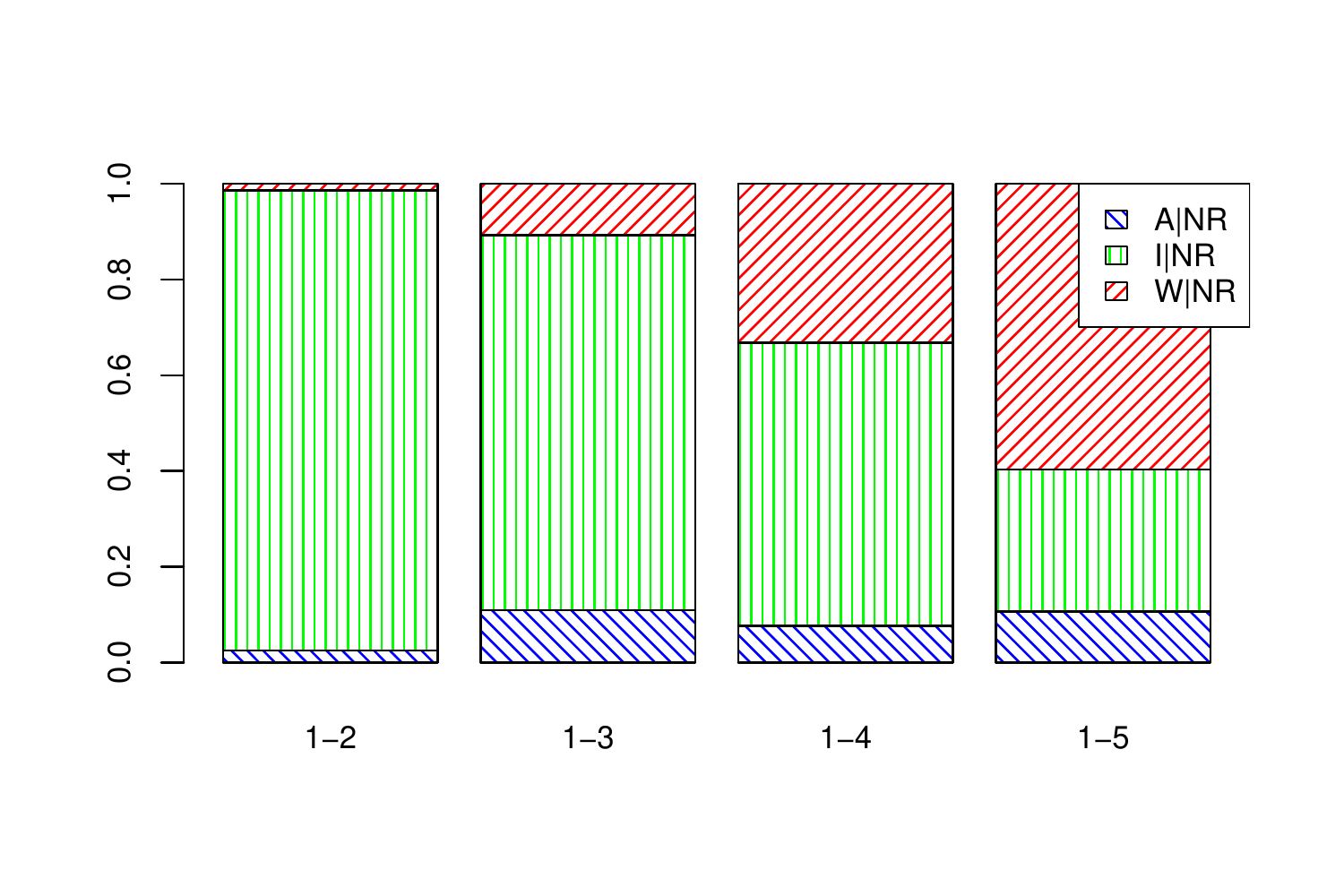}
\vspace{-10mm}
\caption{Knapsack, no-restart, Optimization variant.}
\label{expe:csp-knap-opt}
\end{figure}

\begin{table}[t]
{\scriptsize 
\begin{tabular}{|l|l|r|r|r|r|r|r|}\hline
\multicolumn{2}{|c|}{B} &
\multicolumn{3}{c|}{CSP} &
\multicolumn{3}{c|}{COP} \\
\hline
Bench & Model & $\mu(T)$ & $\sigma(T)$ & $F$ & $\mu(T)$ & $\sigma(T)$ & $F$\\
\hline
1-2 & \abs{}$|$NR & 0.01 & 0.01 & 50 & 0.97 & 0.13 & 50\\
 & \abs$|$R(2) & 0.01 & 0.01 & 50 & 0.74 & 0.08 & 50\\
 & \ibs$|$NR & 0.01 & 0 & 50 & 36.61 & 15.19 & 50\\
 & \ibs$|$R(2) & 0.01 & 0 & 50 & 18.84 & 5.98 & 50\\
 & \wdeg$|$NR & 0 & 0 & 50 & 0.52 & 0.14 & 50\\
 & \wdeg$|$R(2) & 0 & 0 & 50 & 0.60 & 0.11 & 50\\
\hline
1-3 & \abs$|$NR & 0.04 & 0.01 & 50 & 2.03 & 0.27 & 50\\
 & \abs$|$R(2) & 0.04 & 0.01 & 50 & 1.85 & 0.20 & 50\\
 & \ibs$|$NR & 0.02 & 0.01 & 50 & 14.45 & 8.63 & 50\\
 & \ibs$|$R(2) & 0.03 & 0.01 & 50 & 14.01 & 10.15 & 50\\
 & \wdeg$|$NR & 0.01 & 0.01 & 50 & 2 & 0.47 & 50\\
 & \wdeg$|$R(2) & 0.01 & 0.01 & 50 & 2.55 & 0.72 & 50\\
\hline
1-4 & \abs$|$NR & 0.13 & 0.03 & 50 & 26.16 & 7.71 & 50\\
 & \abs$|$R(2) & 0.16 & 0.05 & 50 & 16.35 & 2.11 & 50\\
 & \ibs$|$NR & 0.15 & 0.07 & 50 & 200.96 & 41.91 & 50\\
\hline
\end{tabular}
\begin{tabular}{|l|l|r|r|r|r|r|r|}\hline
\multicolumn{2}{|c|}{Bench} &
\multicolumn{3}{c|}{CSP} &
\multicolumn{3}{c|}{COP} \\
\hline
B & Model & $\mu(T)$ & $\sigma(T)$ & $F$ & $\mu(T)$ & $\sigma(T)$ & $F$\\
\hline
1-4 & \ibs$|$R(2) & 0.20 & 0.1 & 50 & 199.53 & 69.35 & 45\\
 & \wdeg$|$NR & 0.17 & 0.09 & 50 & 112.95 & 33.57 & 50\\
 & \wdeg$|$R(2) & 0.25 & 0.17 & 50 & 195.32 & 36.50 & 48\\
\hline
1-5 & \abs$|$NR & 0.78 & 0.26 & 50 & 53.67 & 13.37 & 50\\
 & \abs$|$R(2) & 0.84 & 0.46 & 50 & 38.68 & 5.26 & 50\\
 & \ibs$|$NR & 2.1 & 1.22 & 50 & 148.89 & 106.74 & 38\\
 & \ibs$|$R(2) & 2.42 & 1.43 & 50 & 101.53 & 83.29 & 45\\
 & \wdeg$|$NR & 1.97 & 0.99 & 50 & 300.01 & 0 & 0\\
 & \wdeg$|$R(2) & 3.98 & 2.12 & 50 & 300.01 & 0 & 0\\
\hline
1-6 & \abs$|$NR & 14.48 & 7.55 & 50 &  &  & \\
 & \abs$|$R(2) & 19.81 & 12.66 & 50 &  &  & \\
 & \ibs$|$NR & 61.57 & 38.66 & 50 &  &  & \\
 & \ibs$|$R(2) & 107.91 & 59 & 49 &  &  & \\
 & \wdeg$|$NR & 233.61 & 81.65 & 28 &  &  & \\
 & \wdeg$|$R(2) & 289.37 & 31.61 & 7 &  &  & \\
\hline
\end{tabular}
}
\vspace{1mm}
\caption{Experimental Results on Multi-Knapsack.}
\label{expe:knap}
\vspace{-1cm}
\end{table}

This benchmark is from \cite{Refalo04} and our implementation follows
Refalo's informal description. The satisfaction model uses an
arithmetic encoding of the binary knapsacks (not a global constraint)
where the objective is replaced by a linear equality with a
right-hand-side set to the known optimal value.  All the constraints
use traditional bound-consistency algorithms for filtering linear
constraints. A second set of experiments considers the optimization
variant.  The COP uses $n$ global binary knapsack constraints
(\texttt{binaryKnapsackAtmost} in \textsc{Comet}) based on the
filtering algorithm in \cite{Trick01adynamic}. These benchmarks
contain up to $50$ variables.

Figure \ref{expe:csp-knap} is a pictorial depiction of the behavior of
the three search algorithms with no restarts. The chart on the left
gives the absolute running time (in seconds) with a logarithmic
scale. The stacked bar chart on the right reports the same data using
a relative scale instead where the height of the bar is the normalized
sum of the running time of all three algorithms and the length of each
segment is its normalized running time. Note that, since this view is
not logarithmic, adjacent bars correspond to different totals.  The
results clearly show that, as the difficulty of the problem increases,
the quality of \wdeg{} sharply decreases and the quality of \abs{}
significantly improves. On the harder instances, \abs{} is clearly
superior to \ibs{} and vastly outperforms \wdeg{}. Figure
\ref{expe:csp-knap-opt} conveys the same information for the
optimization variant with similar observations.

Table \ref{expe:knap} gives the numerical results for instances $1-2$
to $1-6$.  The first column specifies the instance, while the
remaining columns report the average run times, the standard
deviations, and the number of runs that did not time-out. The results
are given for no-restart and slow-restart strategies for all
heuristics. On the decision instance $1-6$, \wdeg{} often fails to
find a solution within the time limit and, in general, takes
considerable time. \abs{} always finds solutions and is about 5 times
faster than \ibs{} for the no-restart strategy which is most effective
on the decision variant. On the optimization variant, \wdeg{} cannot
solve instance $1-5$ in any of the 50 runs and \ibs{} does not always
find a solution. \abs{}, in contrast, finds a solution in all 50 runs
well within the time limit.

In summary, on this benchmark, \wdeg{} is vastly outperformed by
\ibs{} and \abs{} as soon as the instances are not easy. \abs{} is
clearly the most robust heuristic and produces significant
improvements in performance on the most difficult instances, both in
the decision and optimization variants.

\paragraph{Magic Square}

This benchmark is also from \cite{Refalo04} and the model is based on
a direct algebraic encoding with $2\cdot n$ linear equations for the
rows and columns (the square side is $n$), 2 linear equations for the
diagonals, one \texttt{alldifferent} constraint (not enforcing domain
consistency) for the entire square, $2 \cdot n$ binary inequalities to
order the elements in the diagonals, and two binary inequalities to
order the top-left corner against the bottom-left and top-right
corners. Table \ref{expe:magic} report results for squares of size $7$
to size $10$.  The $F$ column in Table \ref{expe:magic} reports the
number of successful runs (no timeout). 

On magic squares, \wdeg{} is completely dominated by \ibs{} and
\abs{}: It has poor performance and is not robust even on the simpler
instances.  The best performance for \ibs{} and \abs{} is obtained
using a fast restart, in which case \abs{} provides a slight
improvement over \ibs{}. \ibs{} is more effective than \abs{} with
slow or no restarts.

\begin{table}[t]
\begin{center}
{\scriptsize
\begin{tabular}{|l|l|r|r|r|r|}
\hline
B & Model & $\mu_C$ & $\mu_T$ & $\sigma_T$ & $F$\\\hline
7 & \abs$|$NR & 8218.06 & 0.53& 1.54& 50\\
 & \abs$|$R(1.1) & 2094.56 & 0.212& 0.12 & 50\\
 & \abs$|$R(2) & 2380.06 & 0.24& 0.11& 50\\
 & \ibs$|$NR & 1030.8 & 0.09& 0.04 & 50\\
 & \ibs$|$R(1.1) & 1172.88 & 0.17& 0.08 & 50\\
 & \ibs$|$R(2) & 961.78 & 0.11& 0.05 & 50\\
 & \wdeg$|$NR & 3294520 & 105.48& 138.24& 34\\
 & \wdeg$|$R(1.1) & 4144754.2 & 146.25& 142.82& 30\\
 & \wdeg$|$R(2) & 218408.26 & 8.03& 42.77& 49\\
\hline
8 & \abs$|$NR & 154783.76 & 7.52& 42.36& 49\\
 & \abs$|$R(1.1) & 5084.18 & 0.48& 0.24& 50\\
 & \abs$|$R(2) & 5941.92 & 0.48& 0.37& 50\\
 & \ibs$|$NR & 1889.4 & 0.21& 0.16& 50\\
 & \ibs$|$R(1.1) & 2694.34 & 0.50& 0.24& 50\\
 & \ibs$|$R(2) & 2524.08 & 0.31& 0.22& 50\\
 & \wdeg$|$NR & 2030330.7 & 79.24& 127.69& 38\\
 & \wdeg$|$R(1.1) & 644467.4 & 28.77& 79.38& 47\\
 & \wdeg$|$R(2) & 339115.4 & 14.96& 59.24& 48\\
\hline
\end{tabular}
\begin{tabular}{|l|l|r|r|r|r|}
\hline
B & Model & $\mu_C$ & $\mu_T$ & $\sigma_T$ & $F$\\\hline
9 & \abs$|$NR & 624461.66 & 37.21& 95.16& 45\\
 & \abs$|$R(1.1) & 12273.72 & 0.96& 0.66& 50\\
 & \abs$|$R(2) & 17552.9 & 1.15& 1.08& 50\\
 & \ibs$|$NR & 630145.78 & 34.19& 91.62& 45\\
 & \ibs$|$R(1.1) & 7239.14 & 1.37& 0.73& 50\\
 & \ibs$|$R(2) & 7622.44 & 0.87& 1.12& 50\\
 & \wdeg$|$NR & 5178690.7 & 243.38& 113.08& 11\\
 & \wdeg$|$R(1.1) & 3588191.4 & 201.01& 126.23& 22\\
 & \wdeg$|$R(2) & 1930318.7 & 96.67& 131.75& 36\\
\hline
10 & \abs$|$NR & 856210.12 & 55.01& 111.94& 42\\
 & \abs$|$R(1.1) & 32404.9 & 2.59& 2.18& 50\\
 & \abs$|$R(2) & 43621.08 & 3.24& 5.04& 50\\
 & \ibs$|$NR & 282909.94 & 18.50& 61.00& 48\\
 & \ibs$|$R(1.1) & 21936.46 & 3.77& 2.90& 50\\
 & \ibs$|$R(2) & 23336.1 & 2.62& 3.04& 50\\
 & \wdeg$|$NR & 4508166.9 & 245.92& 112.18& 10\\ 
 & \wdeg$|$R(1.1) & -& - & - & 0\\
 & \wdeg$|$R(2) & 1825065.5 & 99.70& 125.56& 34\\
\hline
\end{tabular}
}
\end{center}
\caption{Experimental Results on Magic Squares.}
\label{expe:magic}
\end{table}

\paragraph{Progressive Party}

The progressive party problem \cite{Smith96} is a constraint
satisfaction problem featuring a mix of global constraint and has been
used frequently for benchmarking CP, LS, and MIP solvers.  The
instance considered here is the $2-8$ instance with 29 guests, 8
periods and 13 hosts, i.e., 232 variables with domains of size 13.
The goal is to find a schedule for a social event taking place over
$k$ time periods subject to constraints on the sizes of the venues
(the boats), sizes of the group, and social constraints (two groups
cannot meet more than once and one group cannot go back to the same
boat more than once). The model relies on multiple global
alldifferent, multi-knapsacks and arithmetic constraints with
reifications.  This model breaks the search in $k$ phases (one per
period) and uses the black-box heuristic within each period.

\begin{table}[t]
{\scriptsize
\begin{center}
\begin{tabular}[t]{|l|r|r|r|r|}
\hline
Row Labels & $\mu(C)$ & $\mu(T)$ & $\sigma(T)$ & $F$\\\hline
\abs$|$NR & 153848.80 & 46.49 & 90.24 & 45\\
\abs$|$R(1.1) & 2338.18 & 4.91 & 0.87 & 50\\
\abs$|$R(2) & 4324.88 & 5.47 & 2.10 & 50\\
\ibs$|$NR & 1041257.62 & 112.31 & 136.07 & 34\\
\ibs$|$R(1.1) & 15357.24 & 5.24 & 2.23 & 50\\
\hline
\end{tabular}
\begin{tabular}[t]{|l|r|r|r|r|}
\hline
Row Labels & $\mu(C)$ & $\mu(T)$ & $\sigma(T)$ & $F$\\\hline
\ibs$|$R(2) & 60083.46 & 10.03 & 16.89 & 50\\
\wdeg$|$NR & 405027.32 & 93.91 & 128.18 & 37\\
\wdeg$|$R(1.1) & 14424.60 & 3.49 & 2.79 & 50\\
\wdeg$|$R(2) & 19594.12 & 4.00 & 4.50 & 50\\
\hline
\end{tabular}
\end{center}
}
\caption{Experimental Results on the Progressive Party $2-8$.}
\label{expe:pro}
\end{table}

The results are given in Table \ref{expe:pro}. \abs{} clearly
dominates \ibs{} on this benchmark for all restarting
strategies. \abs{} is also clearly superior to \wdeg{} when no
restarts are used but is slightly slower than \wdeg{} when slow or
fast restarts are used.

\paragraph{Nurse Rostering}

This benchmark is taken from \cite{Schaus09} and is a rostering
problem assigning nurses to infants in an hospital ward, while
balancing the workload. The multi-zone model can be found in Listing
1.2 in \cite{Schaus09}. The custom search procedure is removed and
replaced by a call to one of the generic searches
(\ibs{},\abs{},\wdeg{}).  Table \ref{expe:nurse} reports the
performance results for the three heuristics and 3 restarting
strategies on the one-zone instances (z1-z5,z8).  Note that the custom
procedure in \cite{Schaus09} relies on a dynamic-symmetry breaking on
values and sophisticated variable/value ordering. Results for \wdeg{}
beyond \texttt{z5} are not reported as it times out systematically. 
As before, column $F$ reports the number of runs that finish (out of
50), $C$ reports the number of choice points and the $T$ columns
reports the mean and standard deviation of the running time. 

\wdeg{} exhibits extremely poor performance and robustness on this
benchmark.  \abs{} is clearly the most robust procedure as it solves
all instances in all its runs for all the restarting strategies. It is
also significantly faster than \ibs{} on z4 and z8. \ibs{} behaves
well in general, except on z4 and z8 for which it sometimes 
fails to find solutions and takes significantly more time than
\abs{}. It is faster than \abs{} on z3 and z5.

\begin{table}[t]
\begin{center}
{\scriptsize
\begin{tabular}[t]{|l|l|r|r|r|r|r|} \hline
B & Model & $\mu(C)$ & $\mu(T)$ & $\sigma(T)$ & $F$\\\hline
z1 & \abs$|$NR & 282.12 & 0.02 & 0.00 & 50\\
 & \abs$|$R(1.1) & 235.52 & 0.02 & 0.01 & 50\\
 & \abs$|$R(2) & 267.58 & 0.02 & 0.01 & 50\\
 & \ibs$|$NR & 1113.26 & 0.07 & 0.01 & 50\\
 & \ibs$|$R(1.1) & 1028.38 & 0.08 & 0.01 & 50\\
 & \ibs$|$R(2) & 820.52 & 0.07 & 0.01 & 50\\
 & \wdeg$|$NR & 45043.22 & 1.77 & 0.08 & 50\\
 & \wdeg$|$R(1.1) & 63783.44 & 2.46 & 0.17 & 50\\
 & \wdeg$|$R(2) & 47162.36 & 1.87 & 0.08 & 50\\
\hline
z2 & \abs$|$NR & 45223.02 & 2.42 & 0.65 & 50\\
 & \abs$|$R(1.1) & 372174.98 & 19.49 & 9.03 & 50\\
 & \abs$|$R(2) & 98057.72 & 5.03 & 2.53 & 50\\
 & \ibs$|$NR & 82182.32 & 3.84 & 0.91 & 50\\
 & \ibs$|$R(1.1) & 656035.56 & 24.86 & 7.60 & 50\\
 & \ibs$|$R(2) & 177432.42 & 6.78 & 1.96 & 50\\
 & \wdeg$|$NR & 6361685.84 & 300.00 & 0.00 & 1\\
 & \wdeg$|$R(1.1) & 5372380.94 & 300.00 & 0.00 & 3\\
 & \wdeg$|$R(2) & 4944998.26 & 300.00 & 0.00 & 1\\
\hline
z3 & \abs$|$NR & 326902.20 & 23.32 & 10.88 & 50\\
 & \abs$|$R(1.1) & 1944533.10 & 139.55 & 81.15 & 50\\
 & \abs$|$R(2) & 488344.88 & 35.26 & 25.40 & 50\\
 & \ibs$|$NR & 214032.16 & 14.96 & 4.45 & 50\\
 & \ibs$|$R(1.1) & 893297.88 & 62.27 & 12.23 & 50\\
 & \ibs$|$R(2) & 287935.30 & 19.62 & 7.01 & 50\\
\hline
\end{tabular}
\begin{tabular}[t]{|l|l|r|r|r|r|r|} \hline
B & Model & $\mu(C)$ & $\mu(T)$ & $\sigma(T)$ & $F$\\\hline
z3 & \wdeg$|$NR & 4679035.24 & 300.00 & 0.00 & 2\\
 & \wdeg$|$R(1.1) & 5517976.00 & 300.00 & 0.00 & 0\\
 & \wdeg$|$R(2) & 4812533.43 & 300.00 & 0.00 & 2\\
\hline
z4 & \abs$|$NR & 30221.04 & 1.41 & 0.09 & 50\\
 & \abs$|$R(1.1) & 257205.36 & 11.60 & 0.21 & 50\\
 & \abs$|$R(2) & 54855.60 & 2.53 & 0.08 & 50\\
 & \ibs$|$NR & 2794820.62 & 118.25 & 38.18 & 50\\
 & \ibs$|$R(1.1) & 8241949.70 & 300.00 & 0.00 & 2\\
 & \ibs$|$R(2) & 5817294.96 & 221.15 & 42.75 & 47\\
 & \wdeg$|$NR & 6386541.00 & 300.00 & 0.00 & 0\\
 & \wdeg$|$R(1.1) & 5707406.00 & 300.00 & 0.00 & 0\\
 & \wdeg$|$R(2) & 5000897.00 & 300.00 & 0.00 & 0\\
\hline
z5 & \abs$|$NR & 344187.52 & 17.89 & 3.91 & 50\\
 & \abs$|$R(1.1) & 3899344.36 & 185.81 & 38.09 & 50\\
 & \abs$|$R(2) & 902142.38 & 43.40 & 12.82 & 50\\
 & \ibs$|$NR & 120571.44 & 6.16 & 3.61 & 50\\
 & \ibs$|$R(1.1) & 468481.40 & 24.03 & 18.69 & 50\\
 & \ibs$|$R(2) & 192116.32 & 9.84 & 5.27 & 50\\
\hline
z8 & \abs$|$NR & 59314.68 & 3.52 & 0.18 & 50\\
 & \abs$|$R(1.1) & 599777.70 & 36.04 & 3.70 & 50\\
 & \abs$|$R(2) & 119224.04 & 7.00 & 0.53 & 50\\
 & \ibs$|$NR & 7204787.00 & 273.18 & 40.34 & 8\\
 & \ibs$|$R(1.1) & 8201719.08 & 300.00 & 0.00 & 0\\
 & \ibs$|$R(2) & 4301537.08 & 161.61 & 54.71 & 46\\
\hline
\end{tabular}
}
\end{center}
\vspace{-2mm}
\caption{Experimental Results on Nurse Rostering.}
\label{expe:nurse}
\vspace{-5mm}
\end{table}

\paragraph{Radiation}

\begin{table}[t]
\begin{center}
{\scriptsize
\begin{tabular}{l|r}\hline
$B$ & $\sum x \times |D(x)|$ \\\hline
6 & 1x144 + 10x37 + 296x37\\
7 & 1x178 + 10x37 + 333x37\\
8 & 1x149 + 10x37 + 333x37\\ 
9 & 1x175 + 10x37 + 333x37 \\
10 & 1x233 + 10x50 + 441x50 \\
\hline
\end{tabular}
}
\end{center}
\vspace{-2mm}
\caption{Description of the Radiation Instances.}
\label{expe:radspec}
\end{table}

\begin{table}[t]
\begin{center}
{\scriptsize
\begin{tabular}[t]{|l|l|r|r|r|r|}\hline
B & Model & $\mu(C)$ & $\mu(T)$ & $\sigma(T)$ & $F$\\\hline
6 & \abs$|$NR & 14934.94 & 1.99 & 0.65 & 50\\
 & \abs$|$R(1.1) & 10653.36 & 1.49 & 0.39 & 50\\
 & \abs$|$R(2) & 10768.98 & 1.50 & 0.44 & 50\\
 & \ibs$|$NR & 65418.78 & 6.89 & 0.72 & 50\\
 & \ibs$|$R(1.1) & 86200.18 & 8.60 & 0.98 & 50\\
 & \ibs$|$R(2) & 67003.40 & 7.07 & 0.70 & 50\\
 & \wdeg$|$NR & 23279.70 & 1.77 & 0.41 & 50\\
 & \wdeg$|$R(1.1) & 3798.00 & 0.30 & 0.12 & 50\\
 & \wdeg$|$R(2) & 2918.68 & 0.23 & 0.08 & 50\\
\hline
7 & \abs$|$NR & 17434.30 & 2.73 & 1.84 & 50\\
 & \abs$|$R(1.1) & 8481.62 & 1.53 & 0.35 & 50\\
 & \abs$|$R(2) & 9229.80 & 1.62 & 0.51 & 50\\
 & \ibs$|$NR & 90055.32 & 10.42 & 0.44 & 50\\
 & \ibs$|$R(1.1) & 161022.24 & 15.93 & 6.43 & 50\\
 & \ibs$|$R(2) & 98742.94 & 11.13 & 1.73 & 50\\
 & \wdeg$|$NR & 7868.16 & 0.65 & 0.24 & 50\\
 & \wdeg$|$R(1.1) & 2762.26 & 0.24 & 0.10 & 50\\
 & \wdeg$|$R(2) & 2824.00 & 0.24 & 0.12 & 50\\
\hline
8 & \abs$|$NR & 33916.58 & 4.31 & 1.04 & 50\\
 & \abs$|$R(1.1) & 48638.90 & 6.01 & 0.89 & 50\\
 & \abs$|$R(2) & 18294.96 & 2.46 & 0.52 & 50\\
 & \ibs$|$NR & 84329.16 & 8.98 & 1.08 & 50\\
 & \ibs$|$R(1.1) & 187346.80 & 16.94 & 4.97 & 50\\
\hline
\end{tabular}
\begin{tabular}[t]{|l|l|r|r|r|r|}\hline
B & Model & $\mu(C)$ & $\mu(T)$ & $\sigma(T)$ & $F$\\\hline
8 & \ibs$|$R(2) & 88117.48 & 9.36 & 1.34 & 50\\
 & \wdeg$|$NR & 38591.42 & 2.90 & 0.58 & 50\\
 & \wdeg$|$R(1.1) & 20396.80 & 1.72 & 0.39 & 50\\
 & \wdeg$|$R(2) & 6907.14 & 0.55 & 0.12 & 50\\\hline
9 & \abs$|$NR & 40339.62 & 5.79 & 3.36 & 50\\
 & \abs$|$R(1.1) & 20599.88 & 3.21 & 0.35 & 50\\
 & \abs$|$R(2) & 14101.00 & 2.28 & 0.51 & 50\\
 & \ibs$|$NR & 85205.62 & 9.70 & 0.61 & 50\\
 & \ibs$|$R(1.1) & 141311.76 & 14.40 & 3.03 & 50\\
 & \ibs$|$R(2) & 92431.06 & 10.34 & 0.60 & 50\\
 & \wdeg$|$NR & 90489.62 & 7.33 & 1.35 & 50\\
 & \wdeg$|$R(1.1) & 48641.80 & 4.49 & 1.73 & 50\\
 & \wdeg$|$R(2) & 12806.06 & 1.20 & 0.58 & 50\\
\hline
10 & \abs$|$NR & 210181.18 & 34.56 & 17.00 & 50\\
 & \abs$|$R(1.1) & 102777.38 & 17.19 & 3.53 & 50\\
 & \abs$|$R(2) & 50346.82 & 9.10 & 1.65 & 50\\
 & \ibs$|$NR & 2551543.8 & 300.01 & 0.00 & 0\\
 & \ibs$|$R(1.1) & 2504564.1 & 300.01 & 0.00 & 0\\
 & \ibs$|$R(2) & 2525199.8 & 300.01 & 0.00 & 0\\
 & \wdeg$|$NR & 629073.46 & 60.09 & 39.47 & 49\\
 & \wdeg$|$R(1.1) & 232572.16 & 27.88 & 2.28 & 50\\
 & \wdeg$|$R(2) & 47175.04 & 5.60 & 1.30 & 50\\
\hline
\end{tabular}
}
\end{center}
\caption{Experimental Results on Radiation Benchmarks.}
\label{expe:radiation}
\end{table}

This last benchmark is a constrained optimization problem for
\textit{radiation therapy} taken from the 2008 MiniZinc challenge
\cite{Nethercote07minizinc}. The objective is to find a setup of a
radiation therapy machine to deliver a desired radiation intensity to
a tumor. The problem uses algebraic constraint and the formulation can
be found in the mini-zinc repository \cite{mnz-challenge08}\footnote{In this model, 
the time that the beam is on is a variable and must be optimized alongside the number of patterns.}. The
search procedure must deal with all the variables at once. In 2008,
several solvers were unable to solve most instances in a reasonable
amount of time as seen in \cite{mnz-challenge08}, which indicates the
difficulty of the instances.  The instance sizes are specified in
Table \ref{expe:radspec}.  A row gives a term for each array in the
problem with its size and the size of the domains. For instance,
instance 9 has one variable with domain size 175, ten variables with
domain size 37, and 333 variables with domain sizes 37.

Table \ref{expe:radiation} reports the results for 5 instances. \abs{}
clearly dominates \ibs{} on all instances and \ibs{} cannot solve the
largest instance within the time limit for any restarting
strategy. \wdeg{} performs well in general on this benchmark. It is
slightly faster than \abs{} on the largest instance for the slow and
fast restarts, but is slower with no restarts. On instance 9, it is
faster with no restart and slower when slow or fast restart are used. 
Both \wdeg{} and \abs{} are effective on this problem and clearly superior
to \ibs{}. 

\paragraph{Summary} On this collection of benchmarks, \abs{} is
clearly the most robust and effective heuristic. It is robust across
all benchmarks and restarting strategies and is, in general, the
fastest heuristic. \wdeg{} has significant robustness and performance
issues on the multi-knapsack, magic square, and nurse rostering
benchmarks. \ibs{} has some robustness issues on radiation, some
rostering instances, and the optimization variant of the large
knapsack problems. It is in general significantly less efficient than
\abs{} on the knapsack, rostering, and radiation benchmarks.

\subsection{Sensitivity Analysis}

\paragraph{Criticality of the Variable Ordering}

Table \ref{expe:rad-md} reports the performance of activity-based
search when no value ordering is used. The value heuristic simply
tries the value in increasing order as in \wdeg{}.  The results
indicate that the value selection heuristic of \abs{} does not play a critical
role and is only marginally faster/more robust on the largest instances.

\begin{table}[t]
\begin{center}
{\scriptsize
\begin{tabular}[t]{|l|l|r|r|r|r|}\hline
B & Method & $\mu(C)$ & $\mu(T)$ & $\sigma(T)$ & S\\\hline
6 & \abs$|$NR & 11224.80 & 1.48 & 0.58 & 50\\
 & \abs$|$R(1.1) & 18803.18 & 2.30 & 0.86 & 50\\
 & \abs$|$R(2) & 12248.46 & 1.57 & 0.43 & 50\\
\hline
7 & \abs$|$NR & 7147.90 & 1.27 & 0.39 & 50\\
 & \abs$|$R(1.1) & 12161.34 & 1.92 & 0.68 & 50\\
 & \abs$|$R(2) & 10926.12 & 1.74 & 0.54 & 50\\
\hline
8 & \abs$|$NR & 27702.00 & 3.53 & 0.78 & 50\\
 & \abs$|$R(1.1) & 63755.24 & 7.80 & 2.27 & 50\\
 & \abs$|$R(2) & 16708.46 & 2.23 & 0.47 & 50\\
\hline
\end{tabular}
\begin{tabular}[t]{|l|l|r|r|r|r|}\hline
B & Method & $\mu(C)$ & $\mu(T)$ & $\sigma(T)$ & S\\\hline
9 & \abs$|$NR & 36534.92 & 5.06 & 1.18 & 50\\
 & \abs$|$R(1.1) & 46948.84 & 6.76 & 1.99 & 50\\
 & \abs$|$R(2) & 23600.68 & 3.46 & 1.02 & 50\\
\hline
10 & \abs$|$NR & 213094.82 & 33.70 & 9.23 & 50\\
 & \abs$|$R(1.1) & 239145.34 & 40.75 & 7.55 & 50\\
 & \abs$|$R(2) & 87626.36 & 14.87 & 4.14 & 50\\
\hline
\end{tabular}
}
\end{center}
\vspace{-2mm}
\caption{The Influence of the Value Ordering.}
\label{expe:rad-md}
\vspace{-4mm}
\end{table}

\paragraph{Sensitivity to the Sample Size}

\begin{figure}[t]
\includegraphics[width=0.5\columnwidth,trim=20 20 20 20,clip]{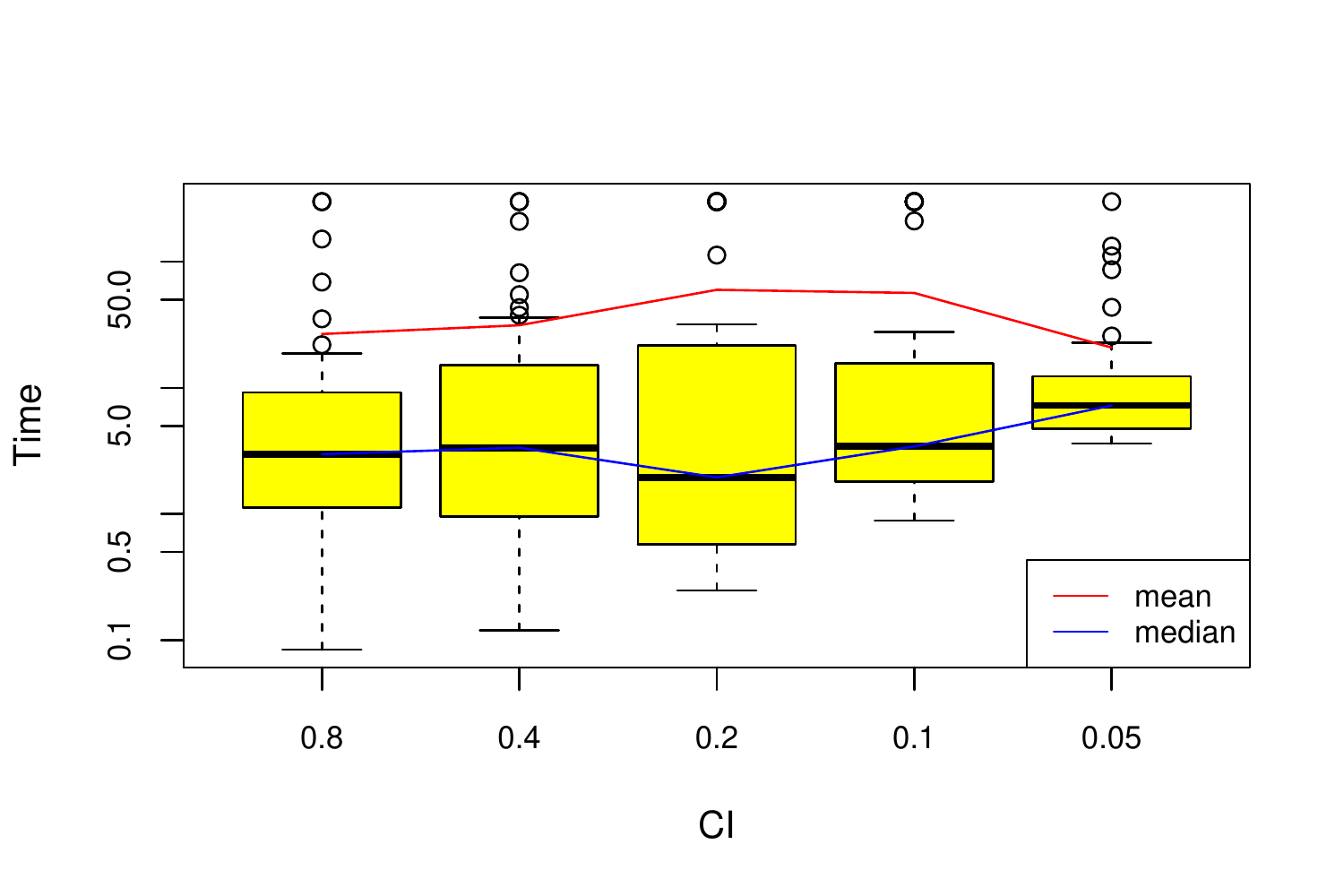}
\includegraphics[width=0.5\columnwidth,trim=20 20 20 20,clip]{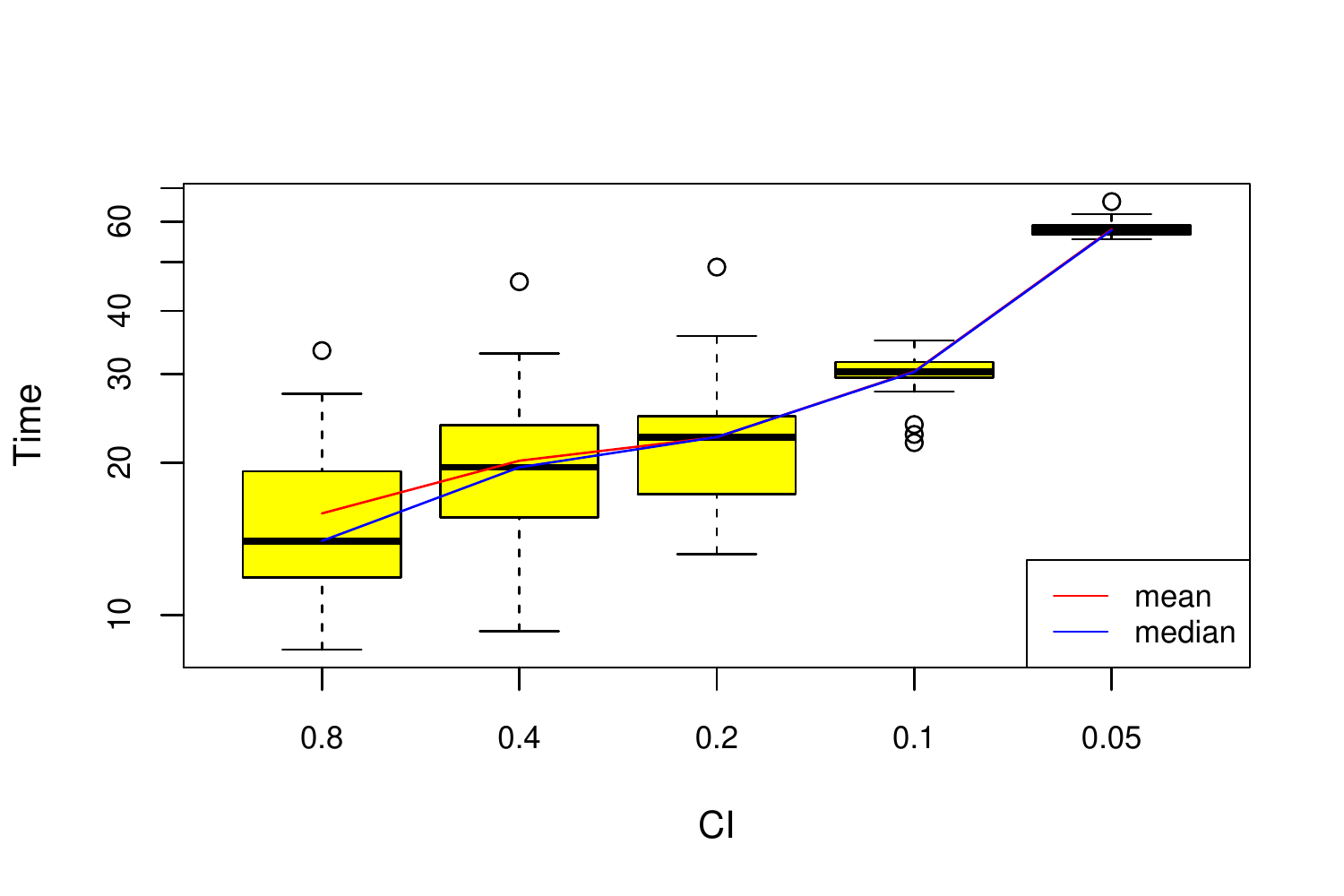}
\caption{Sensitivity to the Sample Size as Specified by $\delta$.}
\label{expe:ci}
\end{figure}

Figure \ref{expe:ci} illustrates graphically the sensitivy of \abs{}
to the confidence interval parameter $\delta$ used to control the
number of probes in the initialization process.  The statistics are
based on 50 runs of the non-restarting strategy. The boxplots show the
four main quartiles for the running time (in seconds) of \abs{} with
$\delta$ ranging from 0.8 down to 0.05.  The blue line connects the
medians whereas the red line connects the means.  The circles beyond
the extreme quartiles are outliers.  The left boxplot shows results on
\texttt{msq-10} while the right one shows results on the optimization
version of \texttt{knap1-4}.

The results show that, as the number of probes increases (i.e.,
$\delta$ becomes smaller), the robustness of the search heuristic
improves and the median and the mean tend to converge. This is
especially true on \texttt{knap1-4}, while \texttt{msq-10} still
exhibits some variance when $\delta=0.05$. Also, the mean decreases
with more probes on \texttt{msq-10}, while the mean increases on
\texttt{knap1-4} as the probe time becomes more important.  The value
$\delta=0.2$ used in the core experiments seem to be a reasonable
compromise. 

\paragraph{Sensitivity to $\gamma$ (Aging)}

\begin{figure}[t]
\includegraphics[width=0.5\columnwidth,trim=20 20 20 20,clip]{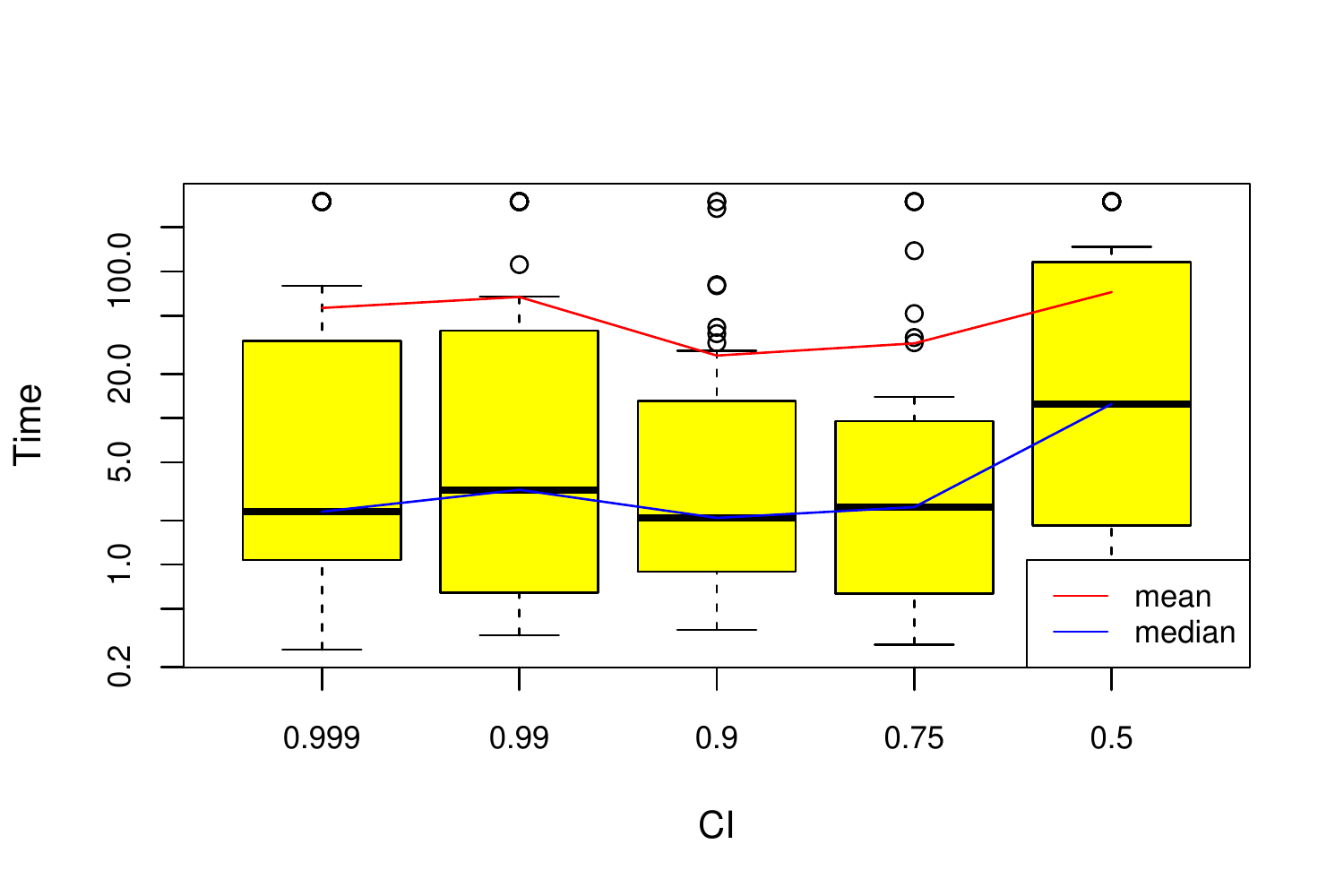}
\includegraphics[width=0.5\columnwidth,trim=20 20 20 20,clip]{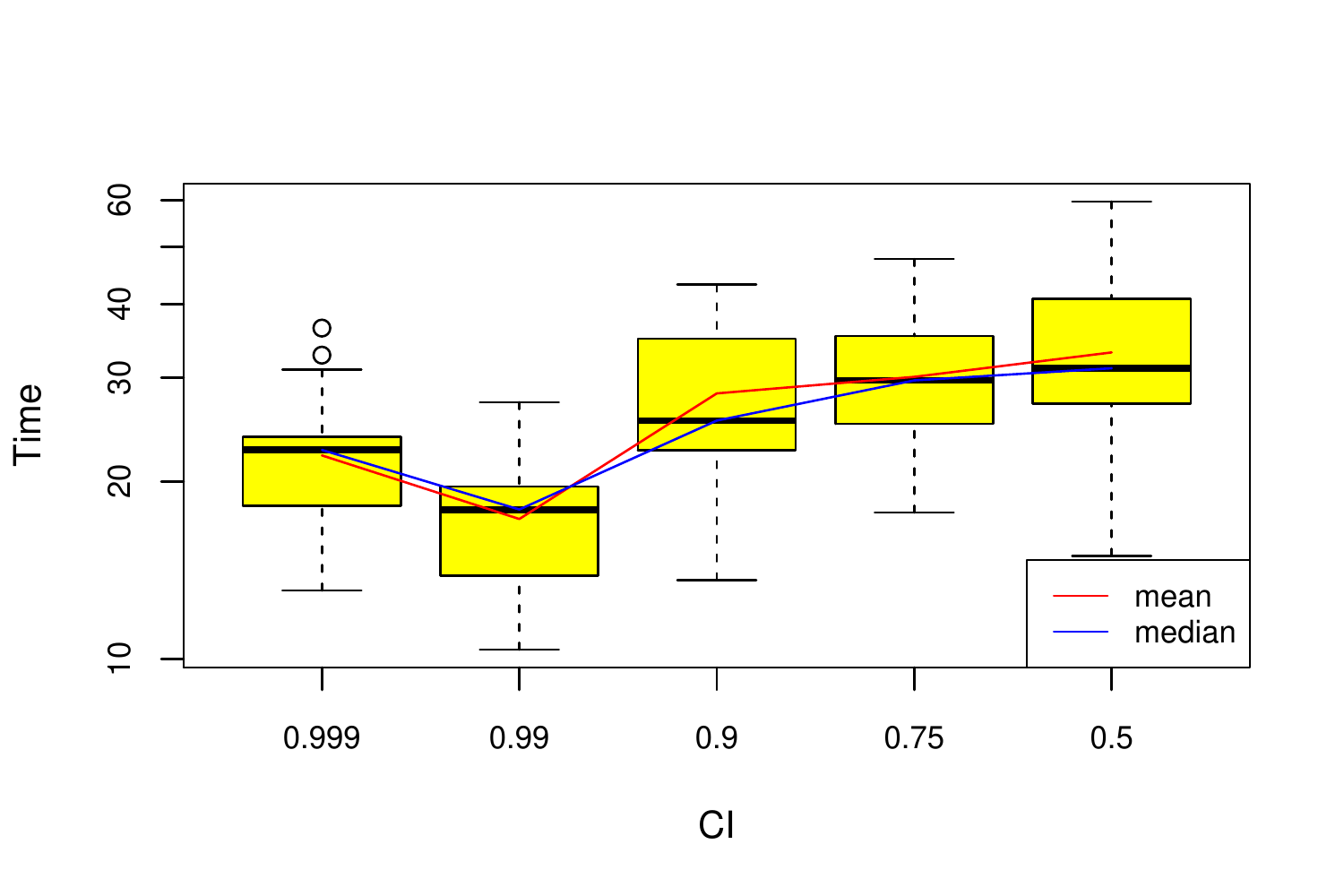}
\caption{Sensitivity to the Aging Process as Specified by $\gamma$.}
\label{expe:gamma}
\end{figure}

Figure \ref{expe:gamma} illustrates the sensitivity to the aging
parameter. The two boxplots are once again showing the distribution of
running times in seconds for 50 runs of \texttt{msq-10} (left) and
\texttt{knap1-4} (right). What is not immediately visible on the
figure is that the number of timeouts for \texttt{msq-10} increases
from $0$ for $\gamma=0.999$ to $9$ for $\gamma=0.5$. Overall, the
results seem to indicate that the slow aging process is desirable.

\subsection{Some Behavioral Observations}

\begin{figure}[t]
\begin{center}
\includegraphics[width=0.55\columnwidth,clip,trim=10 0 20 0]{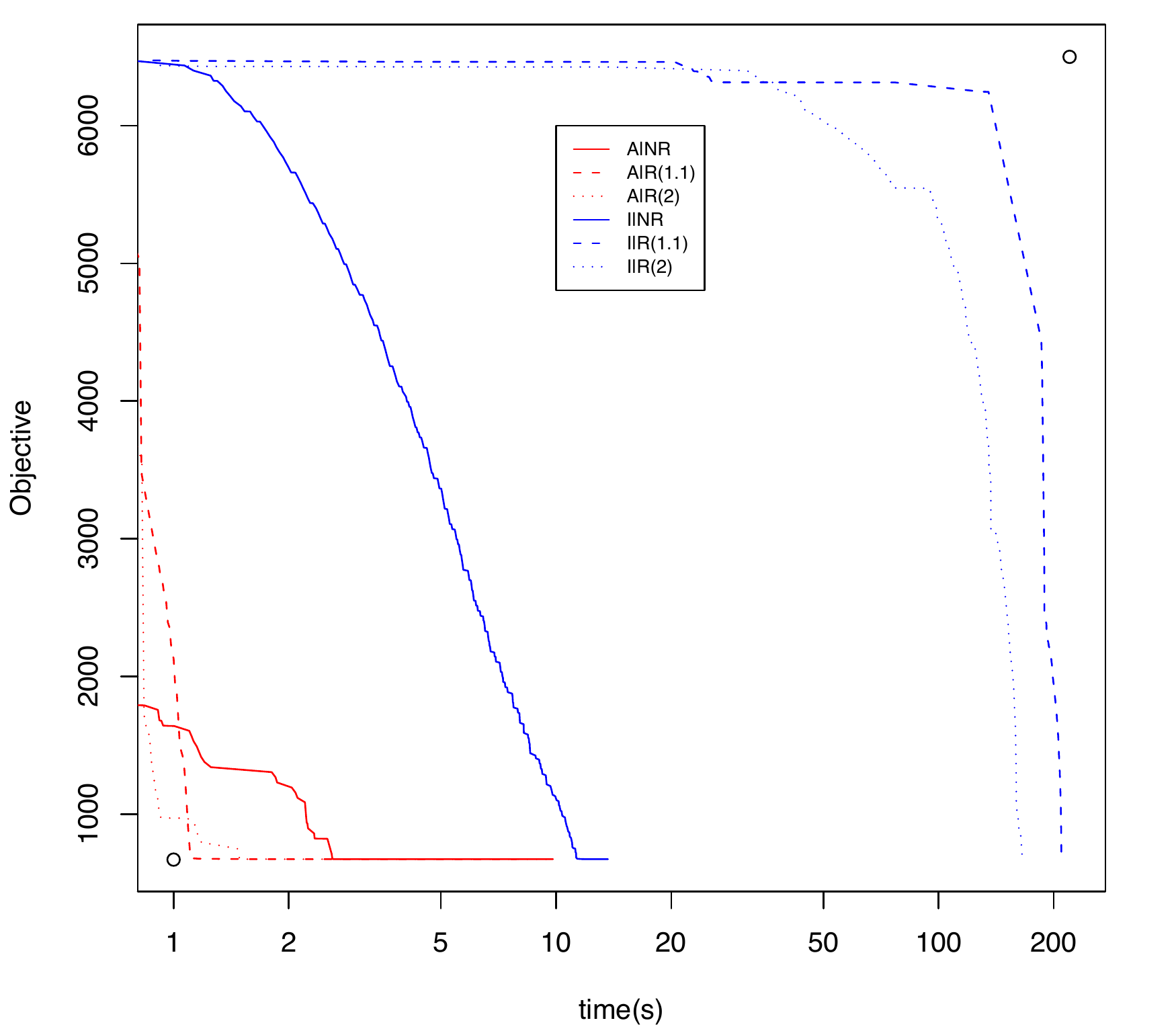}
\end{center}
\vspace{-4mm}
\caption{The Radiation-9 Objective over Time.}
\label{expe:relspeed}
\end{figure}

Figure \ref{expe:relspeed} depicts the behavior of \abs{}  and \ibs{} on  \texttt{radiation \#9}
under all three restarting strategies.  The $x$ axis is the running
time in a logarithmic scale and the $y$ axis is the objective value
each time a new upper bound is found. The three red curves depict the
performance of activity-based search, while the three blue curves
correspond to impact-based search. What is striking here is the
difference in behavior between the two searches. \abs{}
quickly dives to the optimal solution and spends the remaining time
proving optimality. Without restarts, activity-based search hits the
optimum within 3 seconds. With restarts, it finds the optimal solution
within one second and the proof of optimality is faster
too. \ibs{} slowly reaches the optimal solution but then
proves optimality quickly. Restarts have a negative effect on
\ibs{}. We conjecture that the reduction of large domains
may not be such a good indicator of progress and may bias the search
toward certain variables.

\begin{figure}[t]
\includegraphics[width=0.5\columnwidth,trim=20 10 20 20,clip]{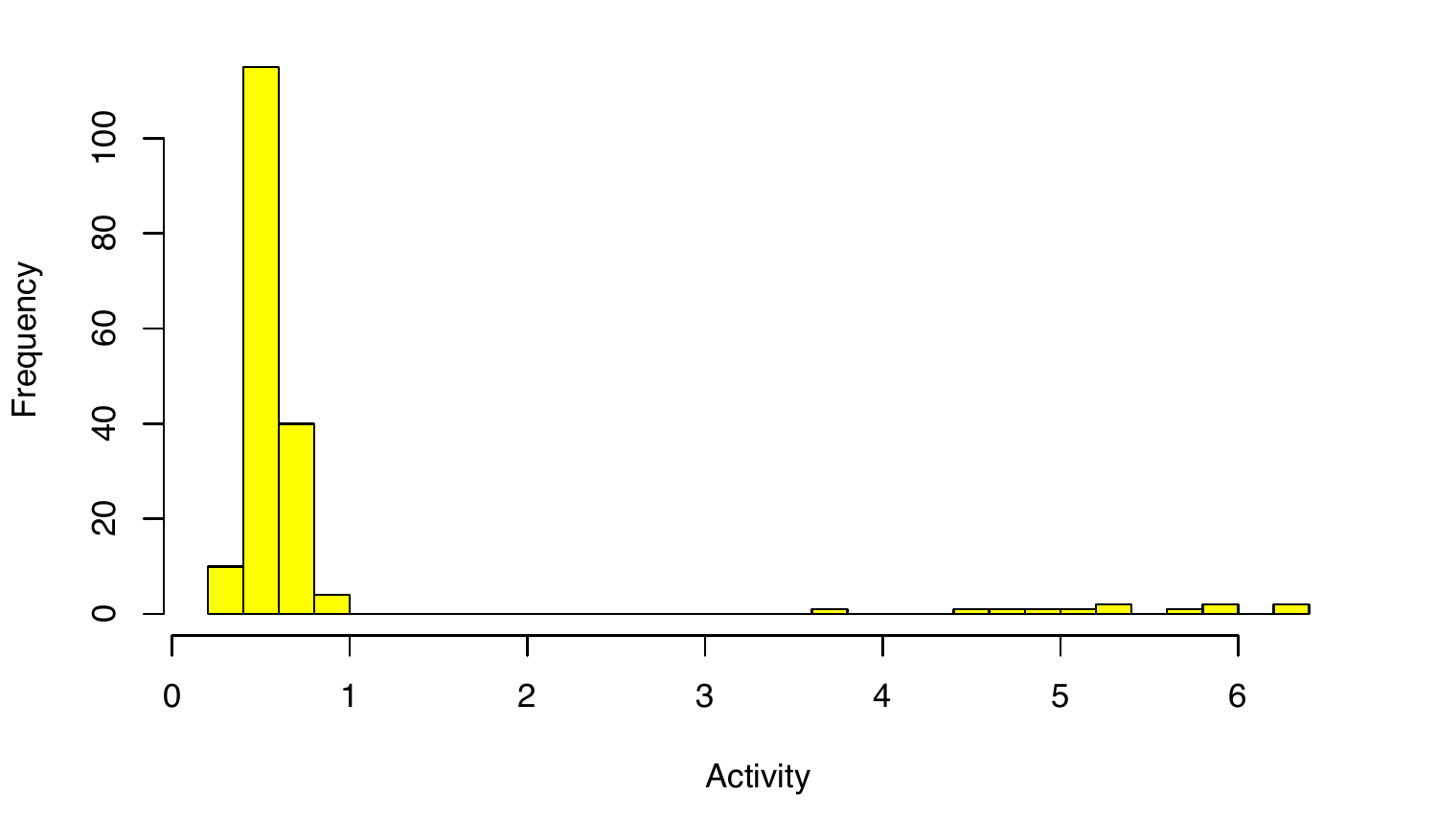}
\includegraphics[width=0.5\columnwidth,trim=20 10 30 20,clip]{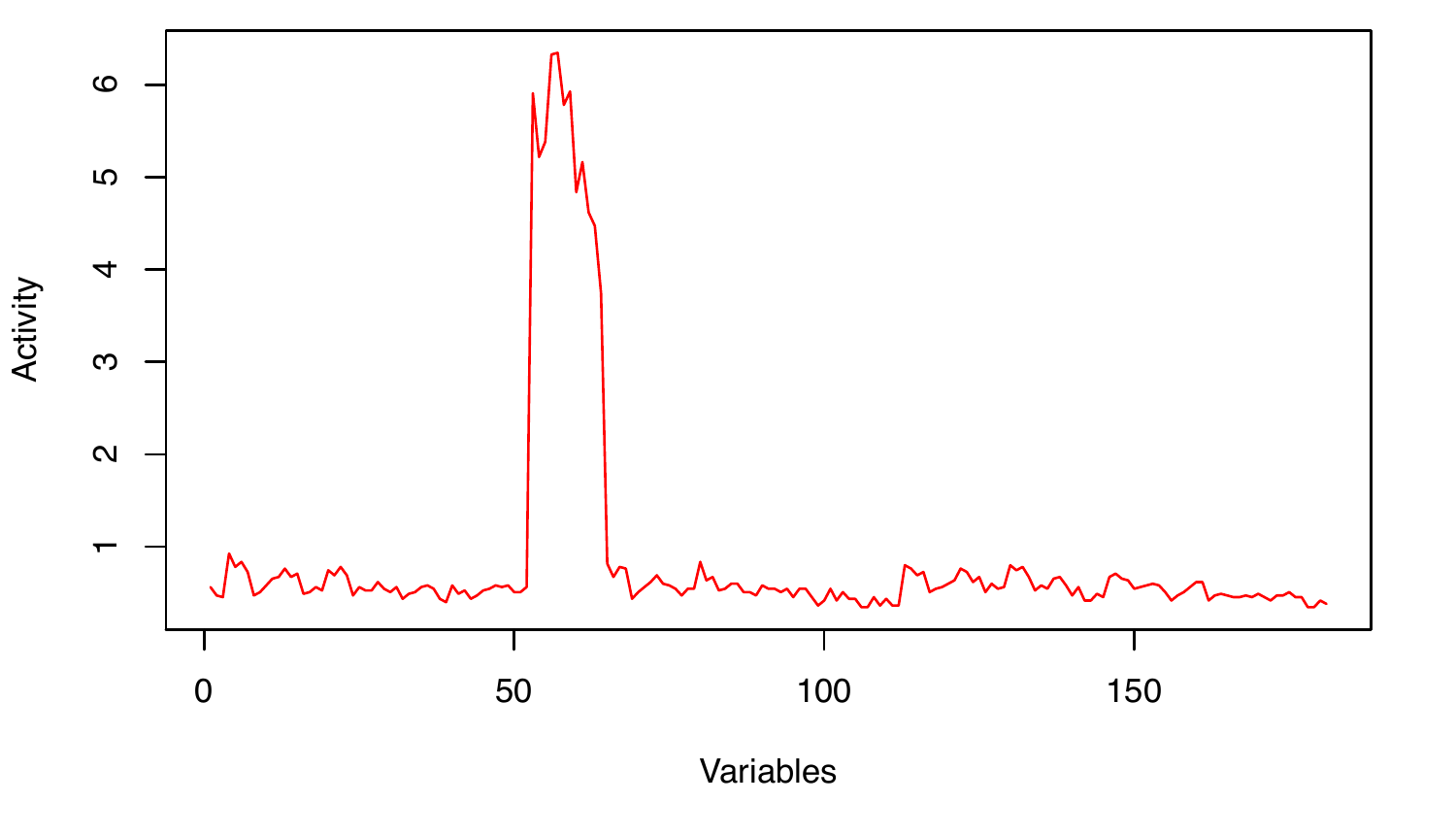}
\caption{Frequencies and Distribution of Activity Levels on Radiation-9.}
\label{expe:histo}
\end{figure}

Figure \ref{expe:histo} provide interesting data about activities on
\texttt{radiation \#9}. In particular, Figure \ref{expe:histo} gives the
frequencies of activity levels at the root, and plots the activity
levels for all the variables. (Only the variables not fixed by
singleton arc-consistency are shown in the figures). The two figures
highlight that the probing process has isolated a small subset of the
variables with very high activity levels, indicating that, on this
benchmark, there are relatively few very active variables. It is
tempting to conjecture that this benchmark has backdoors
\cite{Backdoors03} or good cycle-cutsets \cite{Dechter87} and that
activity-based search was able to discover them, but more experiments
are needed to confirm or disprove this conjecture.

\section{Conclusion}
\label{sec:ccl}

Robust search procedures is a central component in the design of
black-box constraint-programming solvers. This paper proposed
activity-based search, the idea of using the activity of variables
during propagation to guide the search. A variable activity is
incremented every time the propagation step filters its domain and is
aged otherwise. A sampling process is used to initialize the variable
activities prior to search. Activity-based search was compared
experimentally to the \ibs{} and \wdeg{} heuristics on a variety of
benchmarks. The experimental results have shown that \abs{}
was significantly more robust than both \ibs{} and \wdeg{} on these classes
of benchmarks and often produces significant performance improvements.


\begin{thebibliography}{10}
\bibitem{BoussemartHLS04}
F.~Boussemart, F.~Hemery, C.~Lecoutre, and L.~Sais.
\newblock Boosting systematic search by weighting constraints.
\newblock In R.~L. de~M{\'a}ntaras and L.~Saitta, editors, {\em ECAI}, pages
  146--150. IOS Press, 2004.

\bibitem{Dechter87}
R.~Dechter and J.~Pearl.
\newblock The cycle-cutset method for improving search performance in ai
  applications.
\newblock In {\em Proceedings of $3^{rd}$ IEEE conference on AI Applications},
  Orlando, FL, 1987.

\bibitem{Comet10}
I.~Dynadec.
\newblock Comet v2.1 user manual.
\newblock Technical report, Providence, RI, 2009.

\bibitem{mnz-challenge08}
G12.
\newblock \url{http://www.g12.cs.mu.oz.au/minizinc/}, 2008.

\bibitem{Moskewicz:2001}
M.~W. Moskewicz, C.~F. Madigan, Y.~Zhao, L.~Zhang, and S.~Malik.
\newblock Chaff: engineering an efficient sat solver.
\newblock In {\em Proceedings of the 38th annual Design Automation Conference},
  DAC '01, pages 530--535, New York, NY, USA, 2001. ACM.

\bibitem{Nethercote07minizinc}
N.~Nethercote, P.~J. Stuckey, R.~Becket, S.~Brand, G.~J. Duck, and G.~Tack.
\newblock Minizinc: Towards a standard cp modelling language.
\newblock In {\em In: Proc. of 13th International Conference on Principles and
  Practice of Constraint Programming}, pages 529--543. Springer, 2007.

\bibitem{googleSolver10}
L.~Perron.
\newblock \url{http://code.google.com/p/or-tools/}, 2010.

\bibitem{Pesant09}
G.~Pesant.
\newblock Counting and estimating lattice points: Special polytopes for
  branching heuristics in constraint programming.
\newblock {\em Optima Newsletter}, 81:9--14, 2009.

\bibitem{Prosser:2000}
P.~Prosser, K.~Stergiou, and T.~Walsh.
\newblock Singleton consistencies.
\newblock In {\em CP'02}, pages 353--368, London, UK, 2000. Springer-Verlag.

\bibitem{Refalo04}
P.~Refalo.
\newblock Impact-based search strategies for constraint programming.
\newblock In M.~Wallace, editor, {\em CP}, volume 3258 of {\em Lecture Notes in
  Computer Science}, pages 557--571. Springer, 2004.

\bibitem{Schaus09}
P.~Schaus, P.~Van~Hentenryck, and J.-C. RÈgin.
\newblock Scalable load balancing in nurse to patient assignment problems.
\newblock In W.-J. van Hoeve and J.~Hooker, editors, {\em Integration of AI and
  OR Techniques in Constraint Programming for Combinatorial Optimization
  Problems}, volume 5547 of {\em Lecture Notes in Computer Science}, pages
  248--262. Springer Berlin / Heidelberg, 2009.

\bibitem{Smith96}
B.~Smith, S.~Brailsford, P.~Hubbard, and H.~Williams.
\newblock {The Progressive Party Problem: Integer Linear Programming and
  Constraint Programming Compared}.
\newblock {\em Constraints}, 1:119--138, 1996.

\bibitem{Trick01adynamic}
M.~A. Trick.
\newblock A dynamic programming approach for consistency and propagation for
  knapsack constraints.
\newblock In {\em Annals of Operations Research}, pages 113--124, 2001.

\bibitem{Backdoors03}
R.~Williams, C.~P. Gomes, and B.~Selman.
\newblock Backdoors to typical case complexity.
\newblock In {\em Proceedings of the 18th international joint conference on
  Artificial intelligence}, 1173--1178, San Francisco, CA, USA, 2003.
\end{thebibliography}
\end{document}